# TCM-DiffRAG: Personalized Syndrome Differentiation Reasoning Method for Traditional Chinese Medicine based on Knowledge Graph and Chain of Thought

Jianmin Li(1), Ying Chang(3,6), SU-KIT TANG(1), Yujia Liu(3,6), Yanwen Wang(4), Shuyuan Lin(3,6)#, Binkai Ou(2,5)#

#Corresponding author

1. Faculty of Applied Sciences, Macao Polytechnic University, R. de Luís Gonzaga Gomes, 999078, Macao, China.

2. Guangdong Institute of Intelligence Science and Technology, Hengqin, Zhuhai, 519031, Guangdong, China.

3. School of Basic Medical Sciences, Zhejiang Chinese Medical University, 548 Binwen Road, Binjiang District, Hangzhou 310053, China.

4. Hangzhou Ganzhicao Technology Co., Ltd, 10th Floor, Block B, No.11 jugong Road, Binjang District, Hangzhou 310053, China.

5. BoardWare Information System Limited

6. Zhejiang Chinese Medical University – GANCAO DOCTOR Institute of Artificial Intelligence for Chinese Medicine

**Abstract**

**Background:** Retrieval-augmented generation (RAG) technology can empower large language models (LLMs) to generate more accurate, professional, and timely responses without fine-tuning. However, due to the complex reasoning processes and substantial individual differences involved in traditional Chinese medicine (TCM) clinical diagnosis and treatment, traditional RAG methods often exhibit poor performance in this domain.

**Objective:** To address the limitations of conventional RAG approaches in TCM applications, this study aims to develop an improved RAG framework tailored to the characteristics of TCM reasoning.

**Methods:** We developed TCM-DiffRAG, an innovative RAG framework that integrates knowledge graphs (KG) with chains of thought (CoT). TCM-DiffRAG was evaluated on three distinctive TCM test datasets.

**Results:** The experimental results demonstrated that TCM-DiffRAG achieved significant performance improvements over native LLMs. For example, the qwen-plus model achieved scores of 0.927, 0.361, and 0.038, which were significantly enhanced to 0.952, 0.788, and 0.356 with TCM-DiffRAG. The improvements were even more pronounced for non-Chinese LLMs. Additionally, TCM-DiffRAG outperformed directly supervised fine-tuned (SFT) LLMs and other benchmark RAG methods.

**Conclusions:** TCM-DiffRAG shows that integrating structured TCM knowledge graphs with Chain-of-Thought–based reasoning substantially improves performance in individualized diagnostic tasks. The joint use of universal and personalized knowledge graphs enables effective alignment between general knowledge and clinical reasoning. These results highlight the potential of reasoning-aware RAG frameworks for advancing LLM applications in traditional Chinese medicine.

# 1.Introduction

Since the end of 2022, large language models (LLMs) have made remarkable progress, and medical researchers have subsequently used LLMs in various clinical specialties. These preliminary studies[1-4] have shown that LLMs, with their powerful semantic understanding capabilities, exhibit significant advantages over traditional deep learning models in medical natural language processing tasks. However, while general-purpose LLMs have some effectiveness in the medical field, they still fall far short of expert levels[5], and sometimes they may produce incorrect or misleading content due to "hallucinations," which may be related to the quality of training data, model architecture, training processes [6].

Although both fine-tuning [7-14] and RAG [15-25] have been widely practiced in the medical field, they still face many challenges in the field of TCM. Unlike modern medicine, the focus of TCM diagnosis is not on diseases, but on syndromes. A syndrome is a method of classifying pathological symptoms and signs to determine the body's fundamental disorders. TCM obtains a patient's syndrome through syndrome differentiation and treatment. TCM syndrome differentiation theory includes the Eight Principles, Zang-Fu organs, meridians, Qi and blood, or the Sanjiao theory [26]. It analyzes the characteristics of syndromes and diseases based on different parameters such as the function of Zang-Fu organs, the combination of Yin and Yang meridians, and the circulation of Qi and blood [27]. Although the principles of syndrome differentiation and treatment are the same, TCM can be divided into different schools, such as the Classical Formula school, the Earth school, and the Warm Disease school, during the specific diagnosis and treatment process, with certain differences among them [28]. This situation is referred to as "treating different diseases with the same method" and "treating the same disease with different methods."

Previous TCM large language models [14, 29] have done some work in continuous pre-training, supervised fine-tuning, and reinforcement learning. However, due to the lack of diagnostic and treatment data from different TCM schools, it is difficult to reflect the differences in clinical practice. At the same time, the RAG approach also faces numerous challenges in clinical scenarios. TCM clinical problems involve a large amount of potential reasoning. For problems with high complexity and strong logic, embedding models can only match text with surface similarity and cannot identify potential logical structures. In addition, the knowledge base in RAG often comes from textbooks, which is relatively theoretical and has a certain gap from real clinical practice. LLMs find it difficult to effectively derive coherent answers from fragmented or marginally relevant knowledge snippets [30].

To address the challenges of RAG in the context of TCM diagnosis and treatment, recent studies have proposed more advanced RAG methods. For instance, Knowledge Graph RAG transforms the knowledge base into graph structure [31, 32], and Reasoning-based RAG initiates multi-step retrieval to ensure iterative optimization and enhanced reasoning [33]. However, neither of these RAG methods adequately reflects clinical thinking. The core of clinical thinking lies in the dialectical unity of comprehensiveness and focus: it requires not only a high-sensitivity analysis of medical history and symptoms but also targeted, efficient probability assessment and decision-making based on hypotheses [34, 35].

The Knowledge Graph RAG method can manage knowledge hierarchically, connecting different nodes and graphical paths to enhance the logic of recalled texts. However, it demands high-quality graphs and cannot achieve deep iterative reasoning. Reasoning-based RAG can continuously decompose sub-questions, conducting in-depth reasoning retrieval, but the quality of heuristic sub-queries is difficult to control, making it challenging to achieve personalized sub-question decomposition for different TCM schools.

To address these issues, we propose the TCM-DiffRAG method, with the following main contributions:

1. We propose a method for constructing a dual-level knowledge graph specifically for textbooks, further improving the quality of the RAG knowledge base.

2. By combining general knowledge graphs with actual cases, we construct a personalized knowledge graph, compensating for the lack of medical reasoning data and personalized diagnosis and treatment thinking in the knowledge base.

3. We train a TCM thinking chain model, which solves the problems of TCM school adaptation and the unity of retrieval comprehensiveness and focus in RAG practice.

4. We construct a set of evaluation datasets to verify the performance of the TCM thinking chain model and different knowledge bases.

## 2. Methods

### 2.1. Related Work

#### 2.1.1. Retrieval-Augmented Generation

RAG technology requires additional model training and can dynamically integrate external knowledge bases to optimize model output, making it a research hotspot in medical LLM applications [36-38]. The initial RAG, also known as Naive RAG, can be divided into three components: knowledge base, retriever, and large language model. Fig. 1a shows the workflow of Naive RAG, where the user inputs a question, and the retriever (such as the BM25 algorithm or embedding model) retrieves documents from a static dataset. Then, the retrieved documents serve as context to enhance the generation capabilities of the large language model. Naive RAG has some variants, such as Hyde RAG [39, 40]. Early studies focused on improving different RAG components to enhance the effectiveness of RAG [41, 42]. Research has shown that the preprocessing scheme of the knowledge base, the matching degree between the knowledge base and the question, and the capability of the large language model are closely related to the effectiveness of RAG [17, 43]. We have noticed that in some test sets, the performance of stronger models actually decreases after adding RAG [44, 45], which also occurs in our research. This may be because the large model itself has certain medical common sense, and low-quality document recall can negatively impact the capabilities of the large model.

#### 2.1.2. Knowledge Graph and RAG

Knowledge graphs, as structured knowledge bases for representing and linking entities and their relationships in the real world, can play roles such as evidence-based support, link prediction, and flexible multi-hop queries in medical scenarios [46]. Before the rise of LLMs, there were related studies on the combination of knowledge graphs and medical question-answering systems [47-51]. After the rise of LLMs, knowledge graphs, due to their structured semantic associations, provide a new paradigm for deep reasoning of medical data and are often combined with RAG as a knowledge base component. As shown in Fig. 1b, compared to Naive RAG, Knowledge Graph RAG can capture the relationships between entities in the question and recall documents with tighter semantic logic from the knowledge base. In addition, knowledge graphs can perform hierarchical knowledge management, improving the granularity of retrieval [31, 33, 52, 53]. Despite the practical effectiveness of Knowledge Graph RAG, it still faces significant challenges in constructing high-quality knowledge graphs [54], balancing the size of the retrieval subgraph with computational overhead [55], and the lack of multi-step reasoning [56].

#### 2.1.3. Reasoning RAG

Naive RAG cannot decompose complex questions step by step and search for relevant information. Some studies have proposed reasoning-based RAG. Self-BioRAG [57] has the ability to reflect, judge whether the question needs further retrieval, whether the retrieved paragraphs are relevant to the question, and whether the generated content is reasonable. i-MedRAG [43] allows LLMs to iteratively generate subsequent queries, gradually deepening the exploration of information based on previous retrieval results, forming an "information search history," which is ultimately used to generate the final answer. HiRMed [32] also adopts a tree structure, performing medical reasoning at each tree node during RAG. However, most of these studies rely on manually designed prompts or heuristic methods, which not only consume human resources but also lack scalability for more complex problems [58-60]. With the emergence of powerful large reasoning models (LRMs) such as o1and DeepSeek-R1 [61], RAG has begun to enter the era of intelligent agents and deep retrieval. Recent studies [43, 62-64] have integrated the concepts of intelligent agents and reinforcement learning into reasoning-based retrieval. Although reasoning-based RAG can perform in-depth retrieval and reasoning, it still lacks the clinical thinking of a real doctor: the ability to systematically integrate clues to build a holistic picture and the closed-loop reasoning ability to selectively focus on key clues and verify hypotheses.

Based on this, we propose the TCM-DiffRAG method, which combines the advantages of Knowledge Graph RAG and reasoning-based RAG. By training a thinking chain model and constructing a personalized knowledge graph as a knowledge base, we address the adaptation of TCM schools, and the unity of comprehensiveness and focus in retrieval in practice.

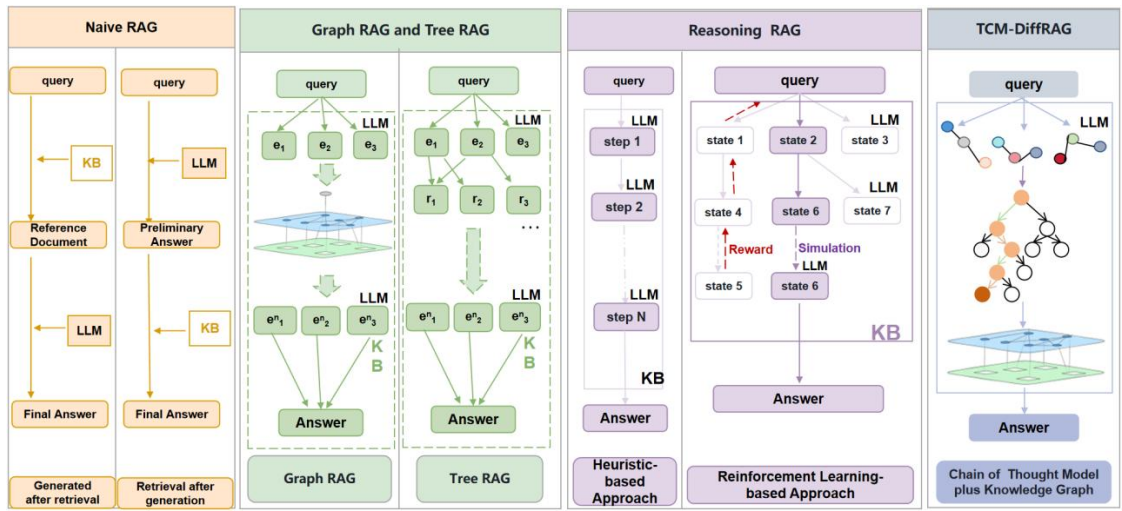

Fig. 1. Comparison of several common RAG methods with the TCM-DiffRAG method.

(Fig. 1a. Naive RAG, which generates answers after only one round of retrieval. Common approaches include generating answers after retrieval or retrieving after generating a preliminary answer.

Fig. 1b. RAG methods using different data structures as knowledge bases. Common approaches include those based on knowledge graphs and those based on thinking trees, which can improve the granularity of retrieval.

Fig. 1c. Reasoning-based RAG. Common approaches include heuristic-based methods and reinforcement learning-based methods. The former relies on the capabilities of large models to gradually decompose complex problems, while the latter incorporates the concepts of intelligent agents and reinforcement learning to train a RAG system that can autonomously optimize retrieval strategies and generation quality.

Fig. 1d. TCM-DiffRAG. Combining the characteristics of Knowledge Graph RAG and reasoning-based RAG, it uses a thinking chain model to decompose complex problems into triples, which are then retrieved and matched with the knowledge graph.)

## 2.2. Methodology

### 2.2.1. Construction of a General Traditional Chinese Medicine Knowledge Graph

The quality of RAG is closely related to the quality of the knowledge base [17, 44]. However, most previous studies on medical knowledge graphs and RAG [31, 52, 53] do not delve deeply into how to construct a knowledge graph. Therefore, we propose a "macro-micro" knowledge graph construction method specifically for medical textbooks. As shown in Fig. 2, we collected 580 classic Chinese medicine textbooks, famous medical cases, and other books, and used previous document parsing research [65-68] for preprocessing. At the macro level of the books, we utilized a document layout model to identify the elements of each PDF page and extracted the titles and corresponding paragraph texts, constructing a knowledge graph similar to a tree diagram. The nodes consist of the titles of the books, and the node relationships are automatically generated through the parent-child structure of the titles. Although this structure sacrifices the traditional ontological concept of knowledge graphs, losing some rigor, it endows therapeutic knowledge with natural semantic indexing capabilities. At the micro level of medical entities, we extracted entities and relationships from paragraph texts using a large language model. The macro title nodes serve as structural hubs, establishing bidirectional mappings with micro entities.

Let the document set D be composed of a chapter hierarchy structure $\mathcal{H}$ and a content collection P:

$$D=\{\mathcal{H},P\}$$

Where $\mathcal{H}$ is a tree-like chapter hierarchy (e.g., Traditional Chinese Medicine Internal Medicine→Chapter 4: Lung System Diseases→Section 2: Cough), and P is the text content corresponding to each chapter. This structure is constructed using a document parsing model. By applying a LLM to D, we extract a set of medical logic triplets $G_{book}$:

$$G_{book}=LLM_{extract}(D)$$

Each triplet $t_k \in G_{book}$ satisfies $t_k=(e_{sub},r,e_{obj})$, where $e_{sub}$ is the subject entity, $r$ is the relation, and $e_{obj}$ is the object entity. There exists a many-to-many mapping between documents and triplets, $\mathcal{M}_1(t_k) \to d_1,d_2,...,d_k$, $\mathcal{M}_1(d_k) \to t_1,t_2,...,t_k$, where $d_1,d_2,...,d_k \in D$.

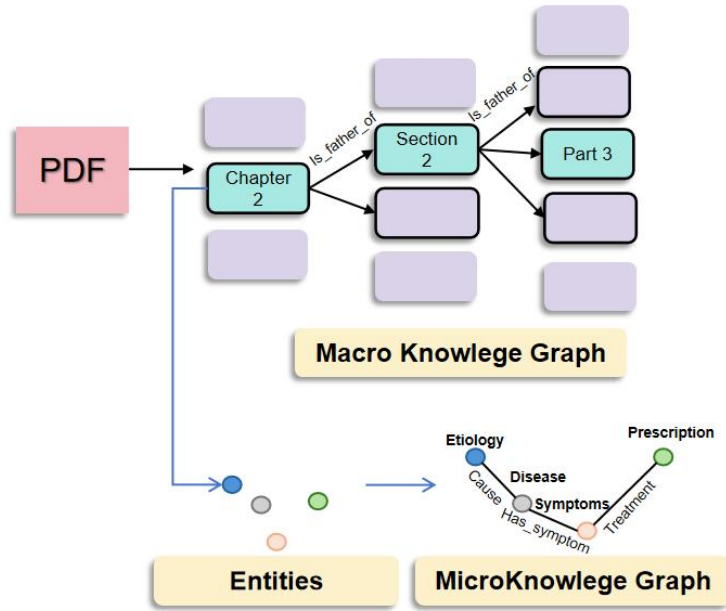

Fig. 2. Schematic Diagram of the General Knowledge Graph Construction.

### 2.2.2. Enhancement and Transfer of the General Traditional Chinese Medicine Knowledge Graph to a Personalized Knowledge Graph

Although we have constructed a general TCM knowledge graph through the study of TCM classics, there are significant differences in clinical diagnosis and treatment among different TCM schools and practitioners. The general TCM knowledge graph and existing RAG solutions cannot effectively capture and reflect these stylistic differences in clinical practice. Integrating diverse clinical thinking patterns into the retrieval and reasoning mechanisms of RAG remains a core challenge. Additionally, the external knowledge base required for complex clinical questions is not explicitly available and needs to be enhanced based on actual cases to improve the matching degree of the RAG knowledge base [69-74]. Drawing inspiration from AgentHospital [75] and MedReason [76], we enhance and transfer the general TCM knowledge graph to obtain a personalized knowledge graph by analyzing the diagnosis and treatment cases of doctors from different schools. The specific method is shown in Fig. 3:

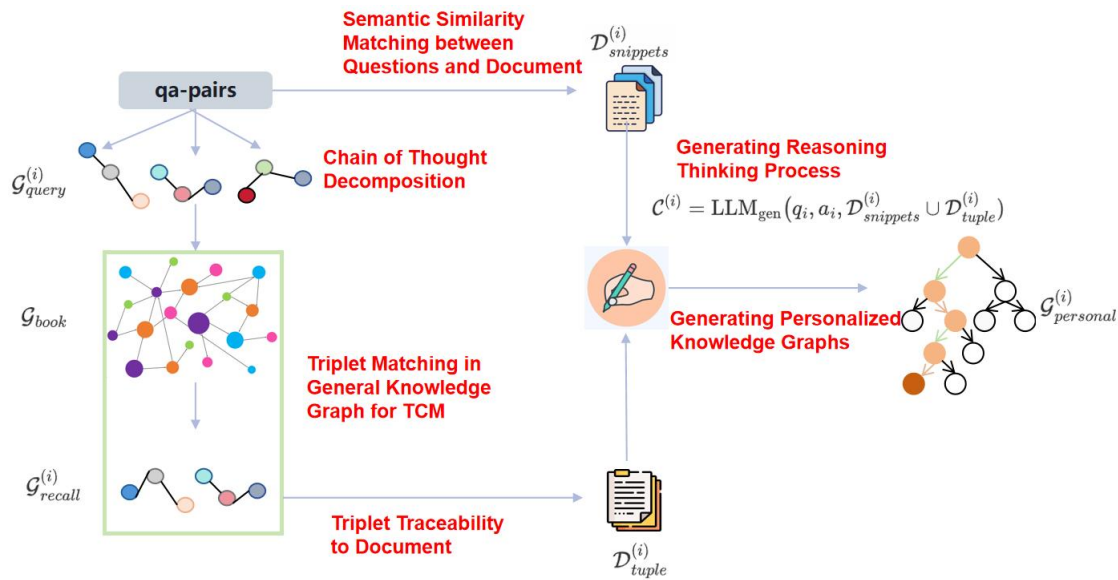

Fig. 3. The Construction Process of the Personalized Knowledge Graph.

**2.2.2.1. Chain of Thought Decomposition**

Input the given question and answer into the Qwen2.5-72B-instruct model to generate a multi-hop reasoning chain and decompose it into structured triplets.

Let Q be the set of questions from different schools of thought, and $A_{gold}$ be the set of their standard answers, derived from the training sets of TCM-MCQ, TCM-SD, and Jingfang-SD datasets. For $q_i \in Q$ and $a_i \in A_{gold}$, generate the general thinking chain triplets:

$$G_{query}^{(i)} = \text{LLM}_{decomp}(q_i, a_i)$$

Where $G_{query}^{(i)}$ is the set of triplets decomposed from the ith question.

**2.2.2.2. Triplet Matching and Traceability**

The generated triplets are aligned with the entities and mapped to the relationships in the general knowledge graph to locate the original text basis in the TCM classics. Firstly, by using vector similarity, retrieve the relevant triplets from the general knowledge graph $G_{book}$, and obtain the recalled set $G_{recall}^{(i)}$:

$$G_{recall}^{(i)} = \underset{t_j \in G_{book}}{\text{argtop-k}} \, \text{sim}\left(\phi(t_j), \phi(G_{query}^{(i)})\right)$$

Where $\phi(\cdot)$ is the embedding operation, and we use Alibaba Cloud's text-embedding-v3. $\text{sim}(\cdot)$ is the cosine similarity. Then, we map $\mathcal{M}_1$ to recall the documents in the TCM classics corresponding to the triplets:

$$D_{tuple}^{(i)} = \bigcup_{t_k \in G_{recall}^{(i)}} \mathcal{M}_1(t_k)$$

#### 2.2.2.3. Questions and Document

Meanwhile, for a given question $q_i$, find the k most relevant text snippets from the document collection D.

$$D_{snippets}^{(i)} = \underset{d_j \in D}{\text{argtop-k}} \; \text{sim}\left(\phi(d_j), \phi(q_i)\right)$$

#### 2.2.2.4. Generating Reasoning Thinking Process

Use the aligned triplets and related classic texts as context to drive the Qwen2.5-72B-instruct model to generate the complete reasoning process from question to answer.

$$C^{(i)} = \text{LLM}_{\text{gen}}(q_i, a_i, D_{snippets}^{(i)} \cup D_{tuple}^{(i)})$$

#### 2.2.2.5. Generating Personalized Knowledge Graphs

Extract new entities and relationships from the reasoning text, integrate them with the original general knowledge graph, and form a knowledge graph containing personalized clinical diagnosis and treatment features. Similar to $G_{book}$, C and $G_{personal}$ have a many-to-many relationship with a mapping relationship $\mathcal{M}_2$ between them, where $\mathcal{M}_2(t_k) \to \{c_1, c_2, \ldots, c_k\}$ and $\mathcal{M}_2(c_k) \to \{t_1, t_2, \ldots, t_k\}$, with $\{c_1, c_2, \ldots, c_k\} \in C$ and $\{t_1, t_2, \ldots, t_k\} \in G_{personal}$.

$$G_{personal}^{(i)} = \text{LLM}_{\text{extract}}(C^{(i)})$$

The advantage of constructing a personalized knowledge graph lies in the decomposition of the reasoning chain through the analysis of doctors' actual diagnosis and treatment cases (Q&A), explicitly capturing their personalized reasoning logic (such as school preferences, emphasis on syndrome differentiation), and avoiding the homogenization defects of the general knowledge graph. At the same time, constrained by the general knowledge graph, while introducing personalized knowledge, it strictly adheres to the core authority of the traditional Chinese medicine theoretical system. More specific examples can be seen in Fig. 4.

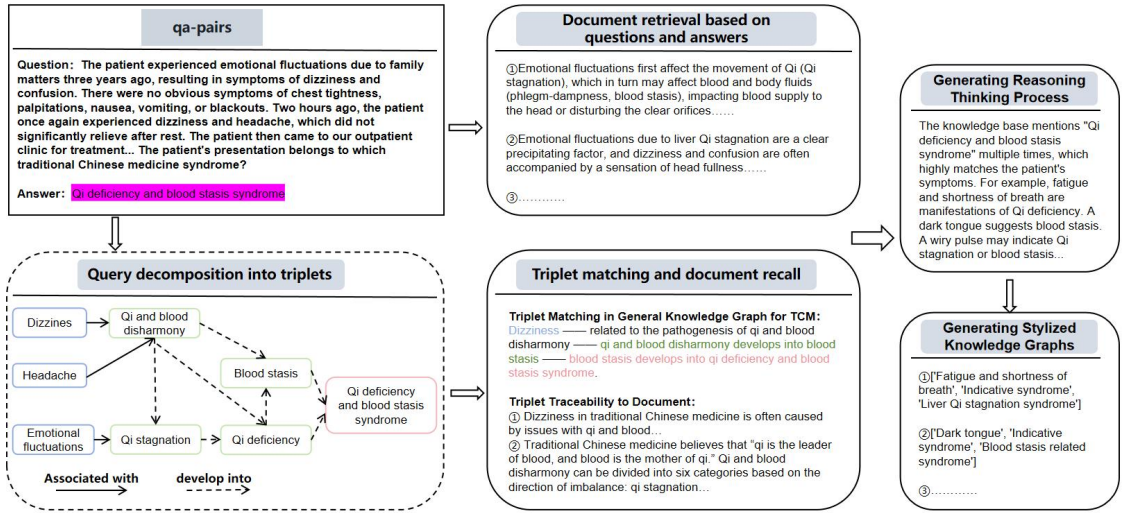

Fig. 4. Specific Examples of the Construction of a Personalized Knowledge Graph.

### 2.2.3. Construction of the TCM-DiffRAG RAG Architecture

After the construction of the personalized knowledge graph, we propose the TCM-DiffRAG Retrieval-Augmented Generation architecture, whose core innovation lies in: decomposing clinical questions into multi-hop triple path sequences through the chain-of-thought reasoning model, and performing semantic alignment and evidence generation based on the personalized knowledge graph. The specific steps are as follows:

#### 2.2.3.1 Chain-of-Thought Model Training

In the previous step, for each question and answer, we generated a dataset C containing question-answer pairs with reasoning processes, as well as the triplets $G_{style}$ obtained by decomposing the reasoning process. These two parts of the data contain a substantial amount of reasoning content, which we use to construct a supervised dataset:

$$D_{SFT} = \{(q_i, C^{(i)})\} \cup \{(q_i, G_{style}^{(i)})\}$$

$\phantom{D_{SFT}=\ }$ Question-answer pair $\phantom{\cup\ }$ Chain of thought triplet

By fine-tuning the large model parameters with domain supervision, we obtain a specialized model $LLM_{cot}$ that possesses the ability to reason about traditional Chinese medicine diagnosis and treatment:

$$\mathscr{L}_{SFT}(\theta) = -\sum_{(q_i, y_i) \in D_{SFT}} \log P_\theta(y_i \mid q_i)$$

We selected the Qwen2.5-7B-instruct model for fine-tuning. The specific training equipment and parameter settings are as follows: We used 8 A800 80G GPUs for training, conducted full parameter fine-tuning based on the LLaMA Factory framework, with a batch size of 2 per GPU, a learning rate of 1e-4, a warm-up ratio of 0.1, and a cutoff_len set to 2046. We used DeepSpeed ZeRO-2 to accelerate training.

#### 2.2.3.2 Multi-hop Retrieval and Knowledge Enhancement

Chain-of-Thought Decomposition: For the input question $q_i$, use $LLM_{cot}$ to parse it into a multi-hop reasoning path.

$$LLM_{cot}(q_i) = \{(s_1, r_1, o_1), \ldots, (s_k, r_k, o_k)\} = T_{query}^{(i)}$$

Personalized Knowledge Recall: The triplets $T_{query}^{(i)}$ from multi-hop reasoning are matched with the personalized knowledge graph $G_{style}$ for semantic similarity, recalling the relevant triplets $T_{recall}^{(i)}$.

$$T_{recall}^{(i)} = \underset{t_j \in G_{style}}{\arg\text{top-}k} \; \text{sim}\left(\phi(t_j), \phi(T_{query}^{(i)})\right)$$

Retrieval of Provisions: $\mathscr{M}_2$ realizes the mapping from triplets to reasoning text provisions C, recalling the relevant text snippets $C_{tuple}^{(i)}$.

$$C_{tuple}^{(i)} = \bigcup_{t_k \in T_{recall}^{(i)}} \mathscr{M}_2(t_k)$$

#### 2.2.3.3 Traceable Diagnosis and Treatment Decision Making

For complex medical questions input by users, the large model generates enhanced responses based on the recalled personalized knowledge graph and its associated classic texts. Thanks to the deep graph traversal capability of the graph (supporting multi-hop reasoning) and the implicit associativity and scalability, the recalled content ensures both the breadth of structured knowledge and the depth of traceable reasoning, thereby integrating the dual advantages of both knowledge graph-based RAG and reasoning-based RAG.

$$\hat{y}_i = \text{LLM}_{\text{gen}}\left(q_i, \underbrace{T_{recall}^{(i)}}_{\text{Reasoning path}}, \underbrace{C_{tuple}^{(i)}}_{\text{Text source}}\right)$$

## 3. Results

### 3.1. Dataset and Knowledge Graph Construction

We divide the corpus dataset into four types: TCM books, TCM-MCQ, TCM-SD, and Jingfang-SD, as shown in Table 1.

The **TCM Books Corpus** constitutes the foundational **General TCM Knowledge Graph**. This corpus includes 580 TCM works, which are divided into macro and micro knowledge graphs through specific methods. The text snippets in the corpus (totaling 433,950) correspond to different hierarchical titles in the books, forming the nodes of the macro knowledge graph. Each text snippet (with an average length of about 330 tokens) represents the specific text content under a certain hierarchical title. We use large language models to extract triple knowledge from each text snippet, with an average of 8 triples extracted per snippet.

The remaining three corpora (TCM-MCQ, TCM-SD, Jingfang-SD) represent different difficulty levels for Retrieval-Augmented Generation (RAG) task evaluation benchmarks:

a. **TCM-MCQ Corpus**: Focuses on testing the mastery of general TCM knowledge. This dataset is derived from TCM medical examination question books, and the task requires selecting the only correct answer from five options. Most of the answer information can be directly retrieved from the TCM books corpus, making this dataset represent the **lowest** task difficulty.

b. **TCM-SD Corpus**: This open-source dataset originates from the real medical records of Xuzhou Hospital of TCM [26]. The task is to determine the only correct answer from 148 candidate syndromes. Compared to TCM-MCQ, the RAG difficulty of this dataset is **significantly increased**, as answers are typically not directly obtainable from the books corpus and require reasoning based on the recalled book snippets.

c. **Jingfang-SD Corpus**: This private dataset comes from the outpatient cases of the Second Affiliated Hospital and the Third Affiliated Hospital of Zhejiang Chinese Medical University. The task requires selecting the only correct answer from 42 candidate syndromes. Its syndrome differentiation approach differs from TCM-SD (based on general principles like the Eight Principles, Zang-Fu organs, meridians, Qi, blood, body fluids, or Sanjiao) and is **rooted in the TCM Classical Formula school, with distinct school characteristics**. Due to the relative scarcity of literature recording such characteristic syndrome differentiation experiences, only a small number of relevant text snippets are available for recall in the books corpus, posing the **greatest challenge** to the RAG system.

The above three evaluation benchmarks (TCM-MCQ, TCM-SD, Jingfang-SD) are divided into training sets and test sets. The training sets are used to construct the personalized knowledge graph, while the test sets are used to evaluate the effectiveness of the RAG system.

Table 1 Composition of the Corpus Database.

| Corpus | Snippets | Average Tokens | Average Triples |
|---|---|---|---|
| TCM Books | 433950 | 330 | 8 |
| TCM-MCQ_Training Set | 21660 | 103 | 12 |
| TCM-MCQ_Test Set | 600 | 117 | / |
| TCM-SD_Training Set | 43085 | 409 | 16 |
| TCM-SD_Test Set | 5486 | 416 | / |
| Jingfang-SD_Training Set | 20049 | 194 | 16 |

| Corpus | Snippets | Average Tokens | Average Triples |
|---|---|---|---|
| Jingfang-SD_Test Set | 5012 | 194 | / |

### 3.2. Evaluation of the General Traditional Chinese Medicine Knowledge Graph Effectiveness

The quality of the corpus, preprocessing, and the method of graph construction have a significant impact on RAG performance. Given that the general Traditional Chinese Medicine (TCM) knowledge graph is the foundational premise for constructing the personalized knowledge graph, it is crucial to evaluate the advancement of this general knowledge graph construction method through experiments. For this purpose, we chose RAGAS [77] as the evaluation framework. This is a tool designed to automate the evaluation of RAG system effectiveness. As shown in Table 2, this study uses OpenAI's gpt-3.5-turbo-16k as the base large language model (LLM), the test set is selected from the TCM-MCQ test set, Alibaba Cloud's text-embedding-v3 is used as the embedding model, and the number of recalled documents is set to k=20. A systematic evaluation was conducted on different corpus processing methods. The experiment compared the following four methods:

a. **Without RAG**: Relying solely on the LLM's own knowledge to generate predicted answers (without retrieval enhancement).

b. **Fixed Character Segmentation**: The TCM books corpus text is segmented into fixed lengths (approximately 330 tokens), resulting in 435,316 document segments. The input questions and document segments are recalled based on semantic similarity after embedding processing.

c. **Macro Knowledge Graph Segmentation**: Segmentation is performed based on the original title hierarchy of the books, generating 433,950 document segments. This method typically produces segments that include titles and their corresponding text content, offering better semantic completeness. Document retrieval is calculated based on cosine similarity:
$D_{snippets}^{(i)} = \underset{d_j \in D}{\text{argtop-k}} \; \text{sim}\left(\phi(d_j), \phi(q_i)\right)$, where $\phi$ represents the embedding function.

d. **Micro Knowledge Graph Segmentation**:

1. First, perform semantic matching retrieval of questions in the triplet set of the general TCM knowledge graph $G_{book}$: $G_{recall}^{(i)} = \underset{t_j \in G_{book}}{\text{argtop-k}} \; \text{sim}\left(\phi(t_j), \phi(q_i)\right)$.

2. Subsequently, retrieve the original text snippets corresponding to the matched triplets based on a predefined **triplet-text snippet mapping** $\mathcal{M}_1$: $D_{tuple}^{(i)} = \cup_{t_k \in G_{recall}^{(i)}} \mathcal{M}_1(t_k)$.

e. **Macro-Micro Knowledge Graph Integrated Retrieval**: Combine the macro-level snippet retrieval $D_{snippets}^{(i)}$ from method (3) with the micro-level triplet-associated snippet retrieval $D_{tuple}^{(i)}$ from method (4), and take their **union** as the final retrieved document set: $D_{final}^{(i)} = D_{snippets}^{(i)} \cup D_{tuple}^{(i)}$.

Table 2  RAGAS Evaluation of the General Traditional Chinese Medicine Knowledge Graph on the TCM-MCQ Test Set.

|  | Accuracy | Answer Similarity | Context Precision | Context Recall | Context Entity Recall |
|---|---|---|---|---|---|
| without RAG | 0.403 | 0.786 | \ | \ | \ |
| Fixed Character Segmentation | 0.540 | 0.856 | 0.621 | 0.829 | 0.173 |

|  | Accuracy | Answer Similarity | Context Precision | Context Recall | Context Entity Recall |
| --- | --- | --- | --- | --- | --- |
| Macro Knowledge Graph Segmentation | 0.640 | 0.863 | 0.808 | 0.848 | 0.188 |
| Micro Knowledge Graph Segmentation | 0.627 | 0.871 | 0.782 | 0.836 | 0.192 |
| Macro-Micro Knowledge Graph Integrated Retrieval | 0.687 | 0.885 | 0.846 | 0.887 | 0.244 |

The experimental results show that the macro-micro knowledge graph integration method leads in all evaluation indicators, verifying the significant advancement of the knowledge graph method we constructed.

Currently, fixed character segmentation is still the method used by most research. To ensure the comparability of document length with subsequent knowledge graph segmentation methods, this study uniformly sets the segmentation at approximately 330 characters. Although this method is simple to implement, it easily leads to semantic fragmentation and entity disconnection. A typical problem is that a large number of tables in textbooks carry key information, and related questions in the test set require complete tables to obtain answers. Fixed character segmentation often truncates tables, destroying their structural integrity and semantic coherence.

In contrast, macro knowledge graph segmentation relies on the original hierarchical structure of books, effectively ensuring the integrity of semantic units. This method performs better when dealing with questions that require cross-document comparison (e.g., what is the preferred treatment plan for low back pain caused by dampness and heat?). However, its retrieval depth has limitations, and its effectiveness may be reduced when the question involves knowledge points buried in text details (e.g., in the Guizhi Decoction and its modified formulas, how does the proportion of Guizhi and Shaoyao change according to clinical conditions?) or when the hierarchical structure causes key entity information to be ineffective.

The advantage of micro knowledge graph segmentation is its ability to relatively accurately match entity relationships. However, it should be noted that the current method directly matches the similarity between questions and triples using vector similarity, lacking an explicit step for extracting key entities in the question, which results in its overall performance being slightly inferior to that of the macro knowledge graph segmentation, showing some improvement only in the Context Entity Recall indicator.
The macro-micro knowledge graph integration method combines the advantages of both: the macro level retains the logical framework and semantic integrity of syndrome differentiation and treatment, while the micro triples precisely lock the core knowledge entities. This synergistic effect enables it to achieve the best comprehensive performance among the four RAG methods.

### 3.3. Evaluation of the Thinking Chain Model's Effectiveness

Using the datasets C and $G_{style}$, which contain the thinking process behind questions, we fine-tuned the qwen-2.5-7B-instruct model to obtain LLM-cot-7B. To evaluate LLM-cot-7B's ability to decompose questions, we selected deepseek-r1 as the reference model to assess the quality of the question-related triples generated by qwen-2.5-7B-instruct and LLM-cot-7B. We used the Likert Scale as the evaluation metric, scoring the triples generated by the two models on a scale from 0 to 5. As shown in the figure, the quality of the triples generated by the fine-tuned LLM-cot-7B is superior to that of qwen-2.5-7B-instruct ($P < 0.05$).

In addition to having better question decomposition capabilities, LLM-cot-7B also has the ability to directly answer questions. The results on the TCM-SD test set show that the effectiveness of LLM-cot-7B (0.74) is significantly better than that of previous studies (0.52) . This confirms that fine-tuning LLMs with data on the thinking process, in addition to the final answer data, can achieve better results. Further ablation experiments indicate that TCM-DiffRAG, using LLM-cot-7B as the thinking chain model in combination with a personalized knowledge graph, outperforms the use of LLM-cot-7B alone in all three test sets.

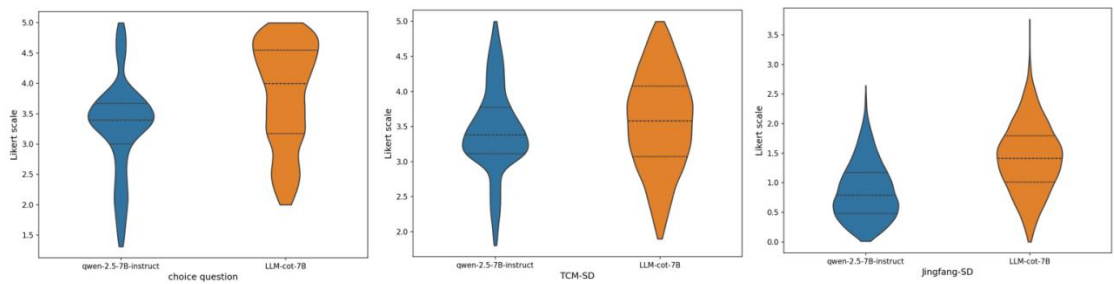

Fig. 5. Comparison of the Quality of Triplet Decomposition by Different Thinking Chain Models.

### 3.4. Evaluation of TCM-DiffRAG's Ablation Experiment

As seen in Figures 6, 7, and 8, when relying solely on the models' own capabilities, qwen-plus and deepseek-r1, which are primarily trained on Chinese language datasets, significantly outperform gpt-4o-mini and gemini-2.5-flash-preview on the TCM-MCQ and TCM-SD tasks. However, on the Jingfangjing-SD test set, all four LLMs perform poorly, indicating that none of them have learned this type of personalized syndrome differentiation thinking during training. Based on the performance of the large models alone, the three test sets represent three different levels of difficulty. The performance of the LLMs is significantly improved when the TCM-DiffRAG method, based on Stylized-KG and LLM-cot-7B, is added.

In the TCM-MCQ test set (Fig. 6), both gpt-4o-mini and gemini-2.5-flash-preview achieve significant improvements with different RAG methods, while qwen-plus and deepseek-r1 only see improvements when using the TCM-DiffRAG method with LLM-cot-7B and a personalized knowledge graph. This may be because qwen-plus and deepseek-r1 already possess excellent general TCM knowledge capabilities, and ordinary RAG methods introduce noisy recalls, leading to negative effects. Only by adding LLM-cot-7B as the thinking chain model and using a more specialized personalized knowledge graph as the knowledge base can these models achieve certain improvements.

The difficulty of the TCM-SD dataset is higher (Fig. 7), and ordinary RAG methods are not effective in improving the performance of LLMs. Significant improvements are observed only after applying the TCM-DiffRAG method. When using LLM-cot-7B as the thinking chain generation model, there is a notable performance improvement compared to the original model. We believe this is because TCM-SD, as a clinical practice test set, requires a high level of reasoning ability from the model. LLM-cot-7B can decompose input queries into finer-grained triples and reason through them. These interconnected triples form a knowledge graph structure with a clinical thinking chain. This method balances the breadth of information retrieval of RAG with the deep reasoning capabilities of the thinking chain.

Jingfang-SD is the most challenging among the three test sets (Fig. 8). Without using RAG, the four generation models can only achieve an accuracy of 0.03-0.07, and there is no significant improvement even with ordinary RAG methods. However, when using LLM-cot-7B in combination with a personalized knowledge graph, the accuracy is increased to 0.35-0.38. The possible reasons for this, in addition to the LLMs' lack of reasoning ability in classical formula diagnosis and treatment, may be the general knowledge graph's lack of classical formula diagnosis and treatment data. More significant improvements can be achieved after using a personalized knowledge graph.

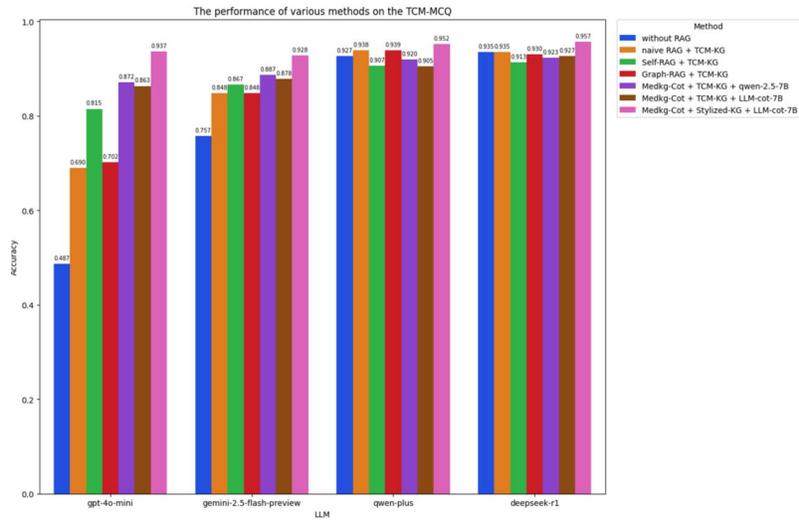

Fig. 6. Performance of Different RAG Methods on the TCM-MCQ Test Set.

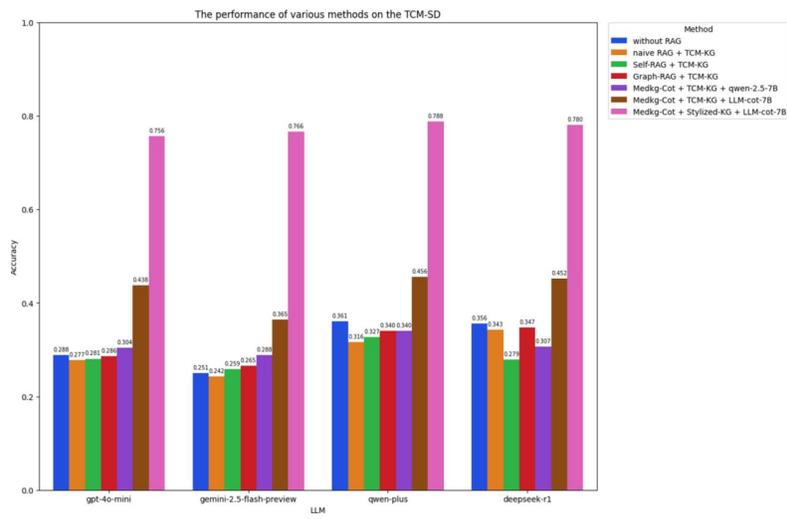

Fig. 7. Performance of Different RAG Methods on the TCM-SD Test Set.

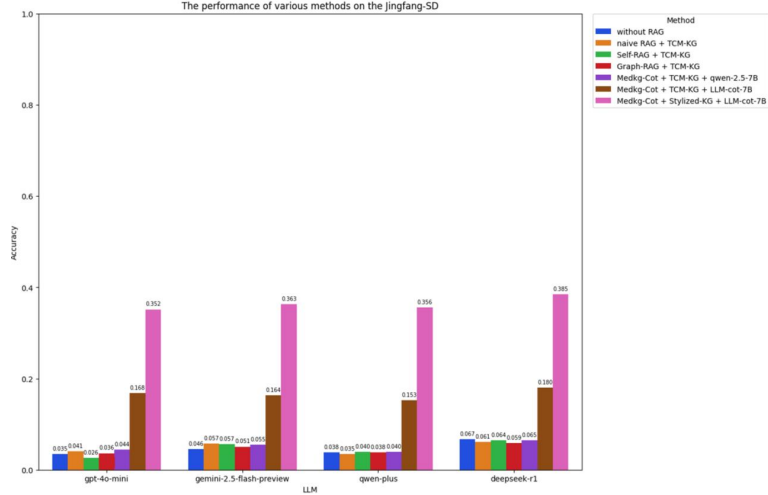

Fig. 8. Performance of Different RAG Methods on the Jingfang-SD Test Set.

## 4.Discussion

This study proposes a RAG method named TCM-DiffRAG, whose core innovation lies in the introduction of a structured knowledge base construction workflow and a training method based on Chain-of-Thought (CoT).

In terms of knowledge base construction, we fully utilize the chapter hierarchy structure of traditional Chinese medicine (TCM) books to segment documents, thereby constructing a universal TCM knowledge graph. Compared to traditional fixed-length character segmentation methods, this semantic unit-based preprocessing approach has significant advantages.

To bridge the gap between general knowledge and specific clinical practice, considering the distinct individualized characteristics of TCM diagnostic thinking, we further build a personalized knowledge graph and its accompanying CoT model based on the aforementioned universal knowledge graph. This CoT model can automatically decompose input questions into multi-hop query triple sequences that conform to a specific style and retrieve subgraphs accordingly. TCM-DiffRAG combines the broad retrieval range advantage of traditional knowledge graph RAGs with the deep reasoning advantage of reasoning-based RAGs.

Ablation test results demonstrate that TCM-DiffRAG, combined with a personalized knowledge graph and CoT model, achieves optimal performance on three benchmark datasets.

Our research conclusions are generalizable and can be extended to other non-medical fields:

**a.TCM-MCQ (Simple Domain Knowledge Question Answering)**: Large language models (LLMs) already perform well on their own in such tasks, and general RAG methods offer limited improvement and may even have a negative effect.

**b. TCM-SD (Domain General Inference Question Answering)**: While LLMs have mastered general industry knowledge, their ability to handle complex reasoning problems remains insufficient. The introduction of a CoT model can significantly enhance the performance of LLMs.

**c. Jingfang-SD (Domain-Specific Inference Question Answering)**: LLMs lack training on such specific private data and perform poorly. This is a typical application scenario for most enterprises, which require reasoning from internal business data. In this scenario, TCM-DiffRAG, combined with a CoT model and personalized knowledge graph, can bring significant performance improvements.

However, this study also has certain limitations, such as: 1. The evaluation dimensions of the knowledge graph need further enrichment and improvement; 2. The performance of TCM-DiffRAG under small sample data conditions requires further research; 3. The differential evaluation of different sizes and bases of LLMs as CoT models remains to be explored. These will be important directions for future research.

**Data availability**

The datasets used and/or analyzed during the current study are available from the corresponding author upon reasonable request.

**Reference**

1. Wilhelm TI, Roos J, Kaczmarczyk R. Large language models for therapy recommendations across 3 clinical specialties: comparative study. J Med Internet Res 2023;25:e49324. [doi:10.2196/49324]

2. Busch F, Hoffmann L, Rueger C, van Dijk EH, Kader R, Ortiz-Prado E, et al. Current applications and challenges in large language models for patient care: a systematic review. Commun Med 2025;5(1):26. [doi:10.1038/s43856-024-00717-2]

3. Zhang K, Meng X, Yan X, Ji J, Liu J, Xu H, et al. Revolutionizing health care: the transformative impact of large language models in medicine. J Med Internet Res 2025;27:e59069. [doi:10.2196/59069]

4. Deng L, Wang T, Yangzhang, Zhai Z, Tao W, Li J, et al. Evaluation of large language models in breast cancer clinical scenarios: a comparative analysis based on chatgpt-3.5, chatgpt-4.0, and claude2. Int J Surg 2024;110(4):1941-1950. [doi:10.1097/JS9.0000000000001066]

5. Hager P, Jungmann F, Holland R, Bhagat K, Hubrecht I, Knauer M, et al. Evaluation and mitigation of the limitations of large language models in clinical decision-making 2024;30(9):2613-2622. [doi:10.1038/s41591-024-03097-1]

6. Ji Z, Lee N, Frieske R, Yu T, Su D, Xu Y, et al. Survey of hallucination in natural language generation. Acm Comput Surv 2023;55(12):248. [doi:10.1145/3571730]

7. Chen Y, Wang Z, Xing X, Zheng H, Xu Z, Fang K, et al. Bianque: balancing the questioning and suggestion ability of health llms with multi-turn health conversations polished by chatgpt. Arxiv E-Prints 2023:2310-15896. [doi:10.48550/arXiv.2310.15896]

8. Li Y, Li Z, Zhang K, Dan R, Jiang S, Zhang Y. Chatdoctor: a medical chat model fine-tuned on a large language model meta-ai (llama) using medical domain knowledge. Cureus 2023;15(6):e40895. [doi:10.7759/cureus.40895]

9. Singhal K, Azizi S, Tu T, Mahdavi SS, Wei J, Chung HW, et al. Large language models encode clinical knowledge. Nature 2023;620(7972):172-180. [doi:10.1038/s41586-023-06291-2]

10. Singhal K, Tu T, Gottweis J, Sayres R, Wulczyn E, Amin M, et al. Toward expert-level medical question answering with large language models. Nat Med 2025;31(3):943-950. [doi:10.1038/s41591-024-03423-7]

11. Wang H, Zhao S, Qiang Z, Li Z, Liu C, Xi N, et al. Knowledge-tuning large language models with structured medical knowledge bases for trustworthy response generation in chinese. Acm Trans Knowl Discov Data 2025;19(2):53. [doi:10.1145/3686807]

12. Yang S, Zhao H, Zhu S, Zhou G, Xu H, Jia Y, et al. Zhongjing: enhancing the chinese medical capabilities of large language model through expert feedback and real-world multi-turn dialogue. Proceedings of the Thirty-Eighth AAAI Conference on Artificial Intelligence and Thirty-Sixth

Conference on Innovative Applications of Artificial Intelligence and Fourteenth Symposium on Educational Advances in Artificial Intelligence: AAAI Press; 2024. p. 2159.

13. Ye Q, Liu J, Chong D, Zhou P, Hua Y, Liu F, et al. Qilin-med: multi-stage knowledge injection advanced medical large language model. Arxiv E-Prints 2023:2310-9089. [doi:10.48550/arXiv.2310.09089]

14. Jia Y, Ji X, Wang X, Zhang H, Meng Z, Zhang J, et al. Qibo: a large language model for traditional chinese medicine. Expert Syst Appl 2025;284:127672. [doi:https://doi.org/10.1016/j.eswa.2025.127672]

15. Xu R, Hong Y, Zhang F, Xu H. Evaluation of the integration of retrieval-augmented generation in large language model for breast cancer nursing care responses. Sci Rep 2024;14(1):30794. [doi:10.1038/s41598-024-81052-3]

16. Ge J, Sun S, Owens J, Galvez V, Gologorskaya O, Lai JC, et al. Development of a liver disease-specific large language model chat interface using retrieval augmented generation. Hepatology 2024(80(5)):1158-1168. [doi:10.1101/2023.11.10.23298364]

17. Kresevic S, Giuffrè M, Ajcevic M, Accardo A, Crocè LS, Shung DL. Optimization of hepatological clinical guidelines interpretation by large language models: a retrieval augmented generation-based framework. Npj Digit Med 2024;7(1):102. [doi:10.1038/s41746-024-01091-y]

18. Liu S, McCoy AB, Wright A. Improving large language model applications in biomedicine with retrieval-augmented generation: a systematic review, meta-analysis, and clinical development guidelines. Journal of the American Medical Informatics Associationjournal of the American Medical Informatics Association 2025;32(4):605-615. [doi:10.1093/jamia/ocaf008]

19. Malik S, Kharel H, Dahiya DS, Ali H, Blaney H, Singh A, et al. Assessing chatgpt4 with and without retrieval-augmented generation in anticoagulation management for gastrointestinal procedures. Ann Gastroenterol 2024;37(5):514-526. [doi:10.20524/aog.2024.0907]

20. Miao J, Thongprayoon C, Suppadungsuk S, Garcia VO, Cheungpasitporn W. Integrating retrieval-augmented generation with large language models in nephrology: advancing practical applications. Medicina (Kaunas) 2024;60(3). [doi:10.3390/medicina60030445]

21. Rau S, Rau A, Nattenmüller J, Fink A, Bamberg F, Reisert M, et al. A retrieval-augmented chatbot based on gpt-4 provides appropriate differential diagnosis in gastrointestinal radiology: a proof of concept study. Eur Radiol Exp 2024;8(1):60. [doi:10.1186/s41747-024-00457-x]

22. Wang D, Liang J, Ye J, Li J, Li J, Zhang Q, et al. Enhancement of the performance of large language models in diabetes education through retrieval-augmented generation: comparative study. J Med Internet Res 2024;26:e58041. [doi:10.2196/58041]

23. Zakka C, Shad R, Chaurasia A, Dalal AR, Kim JL, Moor M, et al. Almanac - retrieval-augmented language models for clinical medicine. Nejm Ai 2024;1(2). [doi:10.1056/aioa2300068]

24. Zelin C, Chung WK, Jeanne M, Zhang G, Weng C. Rare disease diagnosis using knowledge guided retrieval augmentation for chatgpt. J Biomed Inform 2024;157:104702. [doi:10.1016/j.jbi.2024.104702]

25. Woo JJ, Yang AJ, Olsen RJ, Hasan SS, Nawabi DH, Nwachukwu BU, et al. Custom large language models improve accuracy: comparing retrieval augmented generation and artificial intelligence agents to noncustom models for evidence-based medicine. Arthroscopy 2025;41(3):565-573. [doi:10.1016/j.arthro.2024.10.042]

26. Ren M, Huang H, Zhou Y, Cao Q, Bu Y, Gao Y. Tcm-sd: a benchmark for probing syndrome differentiation via natural language processing. Chinese Computational Linguistics: 21st China

National Conference, CCL 2022, Nanchang, China, October 14–16, 2022, Proceedings; Nanchang, China: Springer-Verlag; 2022. p. 247-263.

27. Cao H, Bourchier S, Liu J. Does syndrome differentiation matter? A meta-analysis of randomized controlled trials in cochrane reviews of acupuncture. Med Acupunct 2012;24(2):68-76. [doi:10.1089/acu.2011.0846]

28. Chen K, Xie Y, Liu Y. Profiles of traditional chinese medicine schools. Chin J Integr Med 2012;18(7):534-538. [doi:10.1007/s11655-012-1147-2]

29. Zhang H, Chen J, Jiang F, Yu F, Chen Z, Chen G, et al. Huatuogpt, towards taming language model to be a doctor. Findings of the Association for Computational Linguistics: EMNLP 2023; 1990; Singapore: Association for Computational Linguistics; 2023. p. 10859-10885.

30. Zhao S, Yang Y, Wang Z, He Z, Qiu LK, Qiu L. Retrieval augmented generation (rag) and beyond: a comprehensive survey on how to make your llms use external data more wisely. Arxiv E-Prints 2024:2409-14924. [doi:10.48550/arXiv.2409.14924]

31. Wu J, Zhu J, Qi Y, Chen J, Xu M, Menolascina F, et al. Medical graph rag: evidence-based medical large language model via graph retrieval-augmented generation. Proceedings of the 63rd Annual Meeting of the Association for Computational Linguistics; Vienna, Austria: Association for Computational Linguistics; 2025. p. 28443-28467.

32. Yang Y, Huang C. Tree-based rag-agent recommendation system: a case study in medical test data. Arxiv E-Prints 2025:2501-2727. [doi:10.48550/arXiv.2501.02727]

33. Singh A, Ehtesham A, Kumar S, Talaei Khoei T. Agentic retrieval-augmented generation: a survey on agentic rag. Arxiv E-Prints 2025:2501-9136. [doi:10.48550/arXiv.2501.09136]

34. Joplin-Gonzales P, Rounds L. The essential elements of the clinical reasoning process. Nurse Educ 2022;47(6):E145-E149. [doi:10.1097/NNE.0000000000001202]

35. Yazdani S, Hoseini AM. Five decades of research and theorization on clinical reasoning: a critical review. Adv Med Educ Pract 2019;10:703-716. [doi:10.2147/AMEP.S213492]

36. Lewis P, Perez E, Piktus A, Petroni F, Karpukhin V, Goyal N, et al. Retrieval-augmented generation for knowledge-intensive nlp tasks. Proceedings of the 34th International Conference on Neural Information Processing Systems; Vancouver, BC, Canada: Curran Associates Inc.; 2020. p. 793.

37. Guu K, Lee K, Tung Z, Pasupat P, Chang M. Realm: retrieval-augmented language model pre-training. Proceedings of the 37th International Conference on Machine Learning: JMLR.org; 2020. p. 368.

38. Jiang X, Fang Y, Qiu R, Zhang H, Xu Y, Chen H, et al. Tc–rag: turing–complete rag's case study on medical llm systems. Proceedings of the 63rd Annual Meeting of the Association for Computational Linguistics; Vienna, Austria: Association for Computational Linguistics; 2025. p. 11400-11426.

39. Gao L, Ma X, Lin J, Callan J. Precise zero-shot dense retrieval without relevance labels. Proceedings of the 61st Annual Meeting of the Association for Computational Linguistics (Volume 1: Long Papers); Toronto, Canada: Association for Computational Linguistics; 2023. p. 1762-1777.

40. Jostmann M, Winkelmann H. Evaluation of hypothetical document and query embeddings for information retrieval enhancements in the context of diverse user queries. Wirtschaftsinformatik 2024 Proceedings; 2024. p. 115.

41. Ke YH, Jin L, Elangovan K, Abdullah HR, Liu N, Sia ATH, et al. Retrieval augmented generation for 10 large language models and its generalizability in assessing medical fitness. Npj Digit Med 2025;8(1):187. [doi:10.1038/s41746-025-01519-z]

42. Yang Q, Zuo H, Su R, Su H, Zeng T, Zhou H, et al. Dual retrieving and ranking medical large language model with retrieval augmented generation. Sci Rep 2025;15(1):18062. [doi:10.1038/s41598-025-00724-w]

43. Xiong G, Jin Q, Wang X, Zhang M, Lu Z, Zhang A. Improving retrieval-augmented generation in medicine with iterative follow-up questions. Arxiv E-Prints 2024:2408-2727. [doi:10.48550/arXiv.2408.00727]

44. Xiong G, Jin Q, Lu Z, Zhang A. Benchmarking retrieval-augmented generation for medicine. Findings of the Association for Computational Linguistics: ACL 2024; Bangkok, Thailand: Association for Computational Linguistics; 2024. p. 6233-6251.

45. Vishwanath K, Alyakin A, Alber DA, Lee JV, Kondziolka D, Oermann EK. Medical large language models are easily distracted. Arxiv E-Prints 2025:1201-2504. [doi:10.48550/arXiv.2504.01201]

46. Ebeid IA. Medgraph: a semantic biomedical information retrieval framework using knowledge graph embedding for pubmed. Front Big Data 2022;5:965619. [doi:10.3389/fdata.2022.965619]

47. Li ZQ, Fu ZX, Li WJ, Fan H, Li SN, Wang XM, et al. Prediction of diabetic macular edema using knowledge graph. Diagnostics (Basel) 2023;13(11). [doi:10.3390/diagnostics13111858]

48. Wang L, Xie H, Han W, Yang X, Shi L, Dong J, et al. Construction of a knowledge graph for diabetes complications from expert-reviewed clinical evidences. Comput Assist Surg (Abingdon) 2020;25(1):29-35. [doi:10.1080/24699322.2020.1850866]

49. Zhao X, Wang Y, Li P, Xu J, Sun Y, Qiu M, et al. The construction of a tcm knowledge graph and application of potential knowledge discovery in diabetic kidney disease by integrating diagnosis and treatment guidelines and real-world clinical data. Front Pharmacol 2023;14:1147677. [doi:10.3389/fphar.2023.1147677]

50. Zhou G, E H, Kuang Z, Tan L, Xie X, Li J, et al. Clinical decision support system for hypertension medication based on knowledge graph. Comput Methods Programs Biomed 2022;227:107220. [doi:10.1016/j.cmpb.2022.107220]

51. Lyu K, Tian Y, Shang Y, Zhou T, Yang Z, Liu Q, et al. Causal knowledge graph construction and evaluation for clinical decision support of diabetic nephropathy. J Biomed Inform 2023;139:104298. [doi:10.1016/j.jbi.2023.104298]

52. Zhao X, Liu S, Yang S, Miao C. Medrag: enhancing retrieval-augmented generation with knowledge graph-elicited reasoning for healthcare copilot. Proceedings of the ACM on Web Conference 2025; Sydney NSW, Australia: Association for Computing Machinery; 2025. p. 4442-4457.

53. Chen Q, Ni L. Tcm mlkg-rag: traditional chinese medicine intelligent diagnosis based on multi-layer knowledge graph retrieval-augmented generation. 2024 4th International Conference on Electronic Information Engineering and Computer Communication (EIECC); 2024. p. 958-962.

54. Bui T, Tran O, Nguyen P, Ho B, Nguyen L, Bui T, et al. Cross-data knowledge graph construction for llm-enabled educational question-answering system: a case study at hcmut. Proceedings of the 1st ACM Workshop on AI-Powered Q&A Systems for Multimedia; Phuket, Thailand: Association for Computing Machinery; 2024. p. 36-43.

55. Li M, Miao S, Li P. Simple is effective: the roles of graphs and large language models in knowledge-graph-based retrieval-augmented generation. Arxiv E-Prints 2024:2410-20724. [doi:10.48550/arXiv.2410.20724]

56. Matsumoto N, Moran J, Choi H, Hernandez ME, Venkatesan M, Wang P, et al. Kragen: a knowledge graph-enhanced rag framework for biomedical problem solving using large language models. Bioinformatics 2024;40(6). [doi:10.1093/bioinformatics/btae353]

57. Jeong M, Sohn J, Sung M, Kang J. Improving medical reasoning through retrieval and self-reflection with retrieval-augmented large language models. Bioinformaticsbioinformatics 2024;40(Supplement_1):i119-i129. [doi:10.1093/bioinformatics/btae238]

58. Shao Z, Gong Y, Shen Y, Huang M, Duan N, Chen W. Enhancing retrieval-augmented large language models with iterative retrieval-generation synergy. Findings of the Association for Computational Linguistics: EMNLP 2023; Singapore: Association for Computational Linguistics; 2023. p. 9248-9274.

59. Press O, Zhang M, Min S, Schmidt L, Smith N, Lewis M. Measuring and narrowing the compositionality gap in language models. Findings of the Association for Computational Linguistics: EMNLP 2023; Singapore: Association for Computational Linguistics; 2023. p. 5687-5711.

60. Trivedi H, Balasubramanian N, Khot T, Sabharwal A. Interleaving retrieval with chain-of-thought reasoning for knowledge-intensive multi-step questions. Proceedings of the 61st Annual Meeting of the Association for Computational Linguistics; Toronto, Canada: Association for Computational Linguistics; 2023. p. 10014-10037.

61. DeepSeek-AI, Guo D, Yang D, Zhang H, Song J, Zhang R, et al. Deepseek-r1: incentivizing reasoning capability in llms via reinforcement learning. Arxiv E-Prints 2025:2501-12948. [doi:10.48550/arXiv.2501.12948]

62. Chen M, Li T, Sun H, Zhou Y, Zhu C, Wang H, et al. Research: learning to reason with search for llms via reinforcement learning. Arxiv E-Prints 2025:2503-19470. [doi:10.48550/arXiv.2503.19470]

63. Song H, Jiang J, Min Y, Chen J, Chen Z, Zhao WX, et al. R1-searcher: incentivizing the search capability in llms via reinforcement learning. Arxiv E-Prints 2025:2503-5592. [doi:10.48550/arXiv.2503.05592]

64. Guan X, Zeng J, Meng F, Xin C, Lu Y, Lin H, et al. Deeprag: thinking to retrieve step by step for large language models. Arxiv E-Prints 2025:1142-2502. [doi:10.48550/arXiv.2502.01142]

65. Shen Z, Zhang R, Dell M, Lee BCG, Carlson J, Li W. Layoutparser: a unified toolkit for deep learning based document image analysis. Document Analysis and Recognition – ICDAR 2021: 16th International Conference, Lausanne, Switzerland, September 5–10, 2021, Proceedings, Part I; Lausanne, Switzerland: Springer-Verlag; 2021. p. 131-146.

66. Luo C, Shen Y, Zhu Z, Zheng Q, Yu Z, Yao C. Layoutllm: layout instruction tuning with large language models for document understanding. Arxiv E-Prints 2024:2404-5225. [doi:10.48550/arXiv.2404.05225]

67. Wang B, Xu C, Zhao X, Ouyang L, Wu F, Zhao Z, et al. Mineru: an open-source solution for precise document content extraction. Arxiv E-Prints 2024:2409-18839. [doi:10.48550/arXiv.2409.18839]

68. Blecher L, Cucurull G, Scialom T, Stojnic R. Nougat: neural optical understanding for academic documents. Arxiv E-Prints 2023:2308-13418. [doi:10.48550/arXiv.2308.13418]

69. Zhang T, Madaan A, Gao L, Zheng S, Mishra S, Yang Y, et al. In-context principle learning from mistakes. Arxiv E-Prints 2024:2402-5403. [doi:10.48550/arXiv.2402.05403]

70. Pang C, Cao Y, Ding Q, Luo P. Guideline learning for in-context information extraction. Proceedings of the 2023 Conference on Empirical Methods in Natural Language Processing; Singapore: Association for Computational Linguistics; 2023. p. 15372-15389.

71. Stammer W, Friedrich F, Steinmann D, Brack M, Shindo H, Kersting K. Learning by self-explaining. Arxiv E-Prints 2023:2309-8395. [doi:10.48550/arXiv.2309.08395]

72. Yang L, Yu Z, Zhang T, Cao S, Xu M, Zhang W, et al. Buffer of thoughts: thought-augmented reasoning with large language models. Proceedings of the 38th International Conference on Neural Information Processing Systems; Vancouver, BC, Canada: Curran Associates Inc.; 2025. p. 3607.

73. Zelikman E, Wu Y, Mu J, Goodman ND. Star: self-taught reasoner bootstrapping reasoning with reasoning. Proceedings of the 36th International Conference on Neural Information Processing Systems; New Orleans, LA, USA: Curran Associates Inc.; 2022. p. 1126.

74. Sun H, Jiang Y, Wang B, Hou Y, Zhang Y, Xie P, et al. Retrieved in-context principles from previous mistakes. Proceedings of the 2024 Conference on Empirical Methods in Natural Language Processing; Miami, Florida, USA: Association for Computational Linguistics; 2024. p. 8155-8169.

75. Li J, Wang S, Zhang M, Li W, Lai Y, Kang X, et al. Agent hospital: a simulacrum of hospital with evolvable medical agents. Arxiv Preprint Arxiv:2405.02957 2024. [doi:arXiv:2405.02957]

76. Wu J, Deng W, Li X, Liu S, Mi T, Peng Y, et al. Medreason: eliciting factual medical reasoning steps in llms via knowledge graphs. Arxiv E-Prints 2025:2504-2993. [doi:10.48550/arXiv.2504.00993]

77. Es S, James J, Espinosa Anke L, Schockaert S. Ragas: automated evaluation of retrieval augmented generation. Proceedings of the 18th Conference of the European Chapter of the Association for Computational Linguistics: System Demonstrations; St. Julians, Malta: Association for Computational Linguistics; 2024. p. 150-158.